\documentclass{jdmdh}
\usepackage{array}
\usepackage{pgfplots}
\usepackage{doc}
\usepackage{caption}
\usepackage{booktabs}

\title{Plague Dot Text: Text mining and annotation of outbreak reports of the Third Plague Pandemic (1894-1952)}

\author[1]{Arlene Casey}
\author[2]{Mike Bennett}
\author[1]{Richard Tobin}
\author[1]{Claire Grover}
\author[3]{Iona Walker}
\author[3]{\authorcr Lukas Engelmann}
\author[1,4]{Beatrice Alex}
\affil[1]{Institute for Language, Cognition and Computation, School of Informatics} 
\affil[2]{Digital Library Team, University of Edinburgh Library} 
\affil[3]{Science Technology and Innovation Studies, School of Social and Political Science}
\affil[4]{Edinburgh Futures Institute, School of Literatures, Languages and Cultures}
\affil[ ]{}
\affil[ ]{University of Edinburgh, Edinburgh, UK}

\corrauthor{Beatrice Alex}{balex AT ed.ac.uk}

\begin{document}

\maketitle

\abstract{The design of models that govern diseases in  population is commonly built on information and data gathered from past outbreaks. However, epidemic outbreaks are never captured in statistical data alone but are communicated by narratives, supported by  empirical observations. Outbreak reports discuss correlations between populations, locations and the disease to infer insights into causes, vectors and potential interventions. The problem with these narratives is usually the lack of consistent structure or strong conventions, which prohibit their formal analysis in larger corpora.    Our interdisciplinary research investigates more than 100 reports from the third plague pandemic (1894-1952) evaluating ways of building a corpus to extract and structure this narrative information through text mining and manual annotation.  In this paper we discuss the progress of our ongoing exploratory project, how we enhance optical character recognition (OCR) methods to improve text capture, our approach to structure the narratives and identify relevant entities in the reports. The structured corpus is made available via Solr enabling search and analysis across the whole collection for future research dedicated, for example, to the identification of concepts. We show preliminary visualisations of the characteristics of causation and differences with respect to gender as a result of syntactic-category-dependent corpus statistics. Our goal is to develop structured accounts of some of the most significant concepts that were used to understand the epidemiology of the third plague pandemic around the globe. The corpus enables researchers to analyse the reports collectively allowing for deep insights into the global epidemiological consideration of plague in the early twentieth century.}


%\strut
%\vspace{0.1ex}
%\section{Introduction}
%\label{sect:introduction}
%\strut
%\vspace{5mm}

\section*{Introduction}
\label{sect:introduction}
\vspace{1ex}

The Third Plague Pandemic (1894 - 1950), usually attributed to the outbreak in Hong Kong in 1894, spread along sea trade routes affecting almost every port in the world and almost all inhabited countries, killing millions of people in the late nineteenth and early twentieth centuries \citep{ewing_mapping_2018, echenberg_plague_2007}.  However, as outbreaks differed in severity, mortality and longevity, questions emerged at the time of how to identify the common drivers of the epidemic.  After the Pasteurian Alexandre Yersin had successfully identified the epidemic's pathogen in 1894, \emph{yersinia pestis} \citep{yersin_peste_1894}, the attention of epidemiologists and medical officers turned to the specific local conditions to understand the circumstances by which the presence of plague bacteria turned into an epidemic. These observations were regularly transferred into reports, written to deliver a comprehensive account of the aspects deemed important by the respective author. These reports included discussions, such as extensive elaborations on the social structure of populations, long descriptions of the built environment or close comparison of the outbreak patterns of plague in rats and humans. Many of the reports were quickly circulated globally and served to discuss and compare the underlying patterns and characteristics of a plague outbreak more generally.

These reports are the underlying data set for ongoing work in the \textit{Plague.TXT} project which is conducted by an interdisciplinary team of medical historians, computer scientists and computational linguists. While each historical report was written as a stand-alone document relating to the spread of disease in a particular city, the goal of our work is to bring these reports together as one systematically structured collection of epidemiological reasoning about the third plague pandemic. Given that most reports are already under public domain, this corpus can then be made available to the wider research community through a search interface.  Using methodology from genre analysis \citep{Swales:1990uv}, our approach looks to identify common themes used in the narrative to discuss aspects, such as \textit{conditions, treatments, causes, outbreak history}.  These  themes will then be linked across the report collection.  This allows for comparative analysis across the collection e.g.~comparing discussion on \textit{treatments} or \textit{local conditions}.  In addition to structuring the narrative by theme we also annotate the collection for entities, such as \textit{dates, locations, distances, plague terms}.  This provides for a rich source of information to be tracked and analysed throughout the collection which may unveil interesting patterns with regard to the spread and interpretation of this pandemic.

In the following sections we give an overview of our pilot study firstly describing the background for this project, the data collection, the challenges presented by OCR and improvements we have made to the original digitised reports. Following this we describe our annotation process including our model to structure the reports to extract information. We discuss our combination of manual annotation and automated text mining techniques that support the retrieval and structuring of information from the reports. We discuss our search interface, enabled through Solr, which we use to make the collection available online. Finally, we give some examples of potential use cases for this interface.

\section{Background and Related Work}
\label{sect:background}
\vspace{1ex}

The report collection used in \textit{Plague.TXT} project is a valuable source for multiple historical questions. The pandemic reports offer deep insights into the ways in which epidemiological knowledge about plague was articulated at the time of the pandemic. While they often contain a wealth of statistics and tabulated data, their main value is found in articulated viewpoints about the causes for a plague epidemic, about the attribution of responsibility to populations, locations or climate conditions as well as about evaluating various measurements of control. 

Analysis of reports of the third plague pandemic have been conducted previously, although these centre mainly on manual collation of data using quantitative methods, such as collecting statistics across reports for mortality rates. This derived data has been used to reconstruct transmission trees from localised outbreaks \citep{dean2019}, and to study potential sources and transmission across Europe \citep{bramanti}.  Our \textit{Plague.TXT} project moves beyond existing work by aiming to digitally map epidemiological concepts and themes from the collection of reports, developing pathways to extracting quantitative as well as qualitative information semi-automatically. Combining text mining and manual annotation, we seek to analyse historical plague reports with respect to their narrative  structure.  This allows us to collate  section-specific information, e.g.~\textit{treatments} or discussions on \textit{causes}, for analysis and research.

From the perspective of historiography, this approach also encourages systematical reflections on the underlying conventions of epidemiological writing in the late nineteenth and early twentieth century. Rather than considering reports only within their specific local and historical context, the lateral analysis outlined below contributes to a better understanding of the history of epidemiology as a narrative science \citep{morgan_narrative_2017}. As we engage with the ways in which epidemiologists argued about outbreaks, we identify the concepts they used to investigate the same disease in different locations.  This lateral approach contributes to a better understanding of how these reports were conceived with reference to descriptions and theories used in other reports, and to understanding the epistemological conditions under which epidemiological knowledge began to be formalised at the time on a global scale \citep{morabia_history_2004,ewing_mapping_2018}.

\subsection{Challenges of Understanding Historical Text with Modern Text Mining Tools}
\vspace{1ex}
Whilst there has been a wealth of new tools produced within the text mining community in recent decades, they cannot always be directly used with historical text requiring adaptation for historical corpora.   These changes are due to language evolution resulting not only in differences in style, but also in aspects such as, vocabulary, semantics, morphology, syntax and spelling.   Spelling in historical texts is known to exhibit diachronic, changes over time, and synchronic variance, inconsistencies within the same time period, due to, for example, differences in dialect or spelling conventions \citep{piotrowski2012natural}.  There have been numerous approaches to spelling normalisations, such as those based on rules or edit distances \citep{bollmann2012automatic,pettersson-2013,mitankin2014,pettersson2014}, statistical machine translation \citep{petterson2013,scherrer2013} and more recently neural models \citep{bollmann-etal-2017,korchagina-2017}.  The process of OCR translation of the historical text itself can bring spelling irregularities.  We account for spelling problems in three ways in this work, (cf. Sections \ref{sect:ie}, \ref{sect:dataprep}).  Firstly, we use a lexicon which contains spelling variants found during the pilot annotation when automatically tagging entities, secondly, during the manual annotations any  entities have correct spellings entered. Finally, when inputting our corpus to Solr we correct for spelling mistakes. We say more about general OCR issues in historical text and how we make improvements in processing the OCR generated for our corpus in Section \ref{sec:ocr}.

Differences in syntax can also lead to challenges in using existing NLP tools, such as named entity taggers or part-of-speech taggers \citep{thompson2016}. These rely on accurate identification of syntactic relationships, such as sequences of nouns and adjectives, and word order can be stricter in modern day languages \citep{campbell2013historical,ringe2013}.  Changes of semantic meaning of words provide further challenges, e.g.~widening and narrowing of word senses or change in terminology over time. These can cause issues, such as a reader today may interpret the meaning differently to how it would commonly have been interpreted at the time \citep{Pettersson885117}.   This could also cause problems when searching historical text when different terminology is used for searching, resulting in no results or results that are hard to interpret for the modern-day user.  Using the functionality within Solr we intend to create a map of any such semantic changes to support the challenges that users may encounter when searching. 

\subsection{Understanding the Genre of Outbreak Reports}
\vspace{1ex}
In the reports, outbreaks were described by their occurrence over time.  Statistical data was often  made meaningful through descriptions and theoretical explorations. If the authors  attributed causality, they commonly presented these through careful deliberation of often contradicting hypotheses and theories. Some authors structure their reports using an \textit{introduction, outbreak history, local} or \textit{geographical conditions} followed by discussion on \textit{causes, treatments} and then perhaps a section on \textit{cases}.  Other reports combine this information into sections that discuss all these aspects about a specific location or town and then progress onto a similar discussion about the next town and its localised outbreak. Identifying  the narrative structure is challenging as each report differs in its presentation, ordering and style of content.  

Despite their differences in style, these communicative reports are intended for the same audience of government officials and fellow epidemiologists and will present their arguments comparably.  This study of discourse that shares communicative purpose is called genre analysis \citep{Swales:1990uv}. Recent decades have seen considerable contributions to  understanding how authors structure arguments within specific genres and have demonstrated pathways of how  text mining can be applied to automatically recognise these structures.  Most relevant to our work is that done for scientific articles, such as Argument Zoning \citep{teufel1999argumentative} or Core Scientific Concepts \citep{Liakata:2012fv}.  These works seek to model the intentional structure of a research article.  However, the models proposed are designed to extract different information from different disciplines. For example, Argument Zoning is originally designed for use within the discipline of Computational Linguistics.  When this model is applied within the domain of Chemistry \citep{teufel:2009} it is required to extend it to adequately address aspects of communication that occur within this domain. Although other work that models communicative purpose may have similar goals to ours, it does not adequately represent our needs and does not capture all the aspects of our type of discourse. Hence, we had to develop our own model to capture the information and arguments made within our collection of reports.

Using genre analysis as our methodological approach, treating each report as a communicative event about a specific plague outbreak, our focus is on building a structure to label the information contained in individual reports, such that similar discourse segments can be linked and studied across the collection.  We seek to collate the concepts, themes and approaches  across the report collection to consider comparable conventions within the entire corpus.  For example, we seek to enable comparative analysis of \textit{causes, treatments, local conditions} between different outbreaks, and various times and places.  While there has been previous work on bootstrapping and mapping concepts in other types of historical texts (e.g.~commodities in historical collections on nineteenth century trade in the British empire \citep{klein:2014, hinrichs2015, Clifford:2016}) we are unaware of other work where this is done with respect to narrative document structure across a historical collection.

\subsection{Contribution}
\vspace{1ex}
Our contribution is the development of a systematically structured corpus, which we capture through annotation, to assimilate similar discourse segments such as \textit{causes} or \textit{treatments} across the reports. In addition, we develop an interactive search interface to our collection. 

This search tool in combination with our structure model allows follow-on research to conduct automated or semi-automated exploration of a rich source about the conceptual thinking at the time of the third plague pandemic.  This allows for better understanding of the historical epistemology of epidemiology and to thus provide valuable lessons about dealing with contemporary global spread of disease. The corpus thus constitutes an archive, from which future analysis will discern \emph{concepts}, with which plague has been shaped into an object of knowledge in modern epidemiology. This will also enable new perspectives on the formalisation of epidemiology as a discipline in the twentieth century.

 \begin{figure}[ht]
 \begin{centering}
		\includegraphics[width=0.7\textwidth]{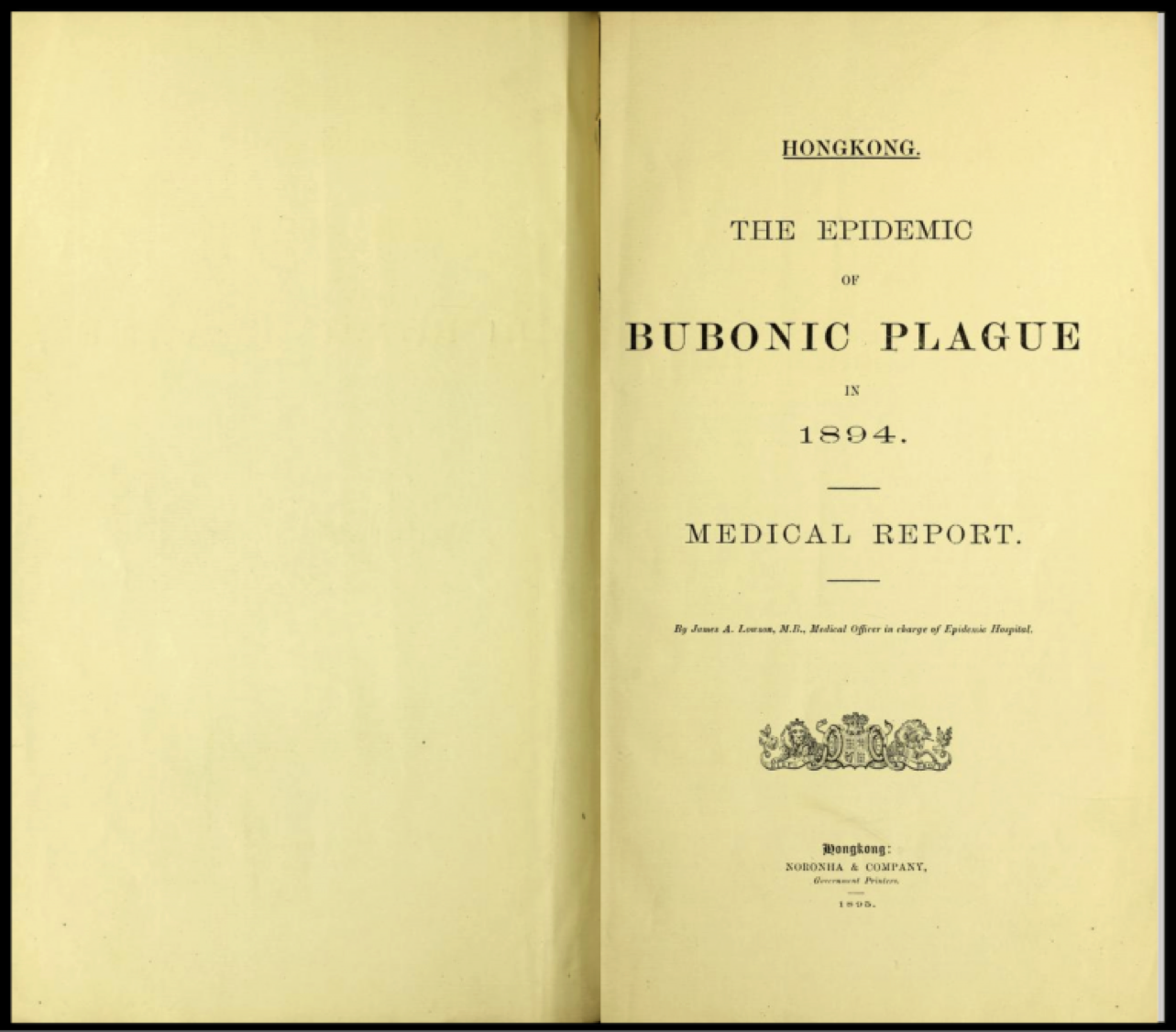}
	\captionof{figure}{Bubonic plague report for Hong Kong \citep{lowson1895epidemic}.}
	\label{fig:report}
 	\end{centering}
 \end{figure}

\section{Data}
\label{sect:data}
\vspace{1ex}

The third plague pandemic was documented in over 100 outbreak reports for most major cities around the world. Many of them have been digitised, converted to text via OCR and are available via the Internet Archive\footnote{\url{https://archive.org/details/b24398287}} and the UK Medical Heritage Library.\footnote{\url{https://wellcomelibrary.org/collections/digital-collections/uk-medical-heritage-library/}}  Figure~\ref{fig:report} shows an example of such a report covering the Hong Kong outbreak which was published in 1895 and is accessible with open access on Internet Archive.

We treat all relevant reports for which we have a scan as one collection.  While the majority of reports in this set (102) are written in English, there are further reports in French, Spanish, Portuguese and other languages which we excluded from the analysis at this stage.

 \begin{figure}[ht]
 \begin{centering}
		\includegraphics[width=0.7\textwidth]{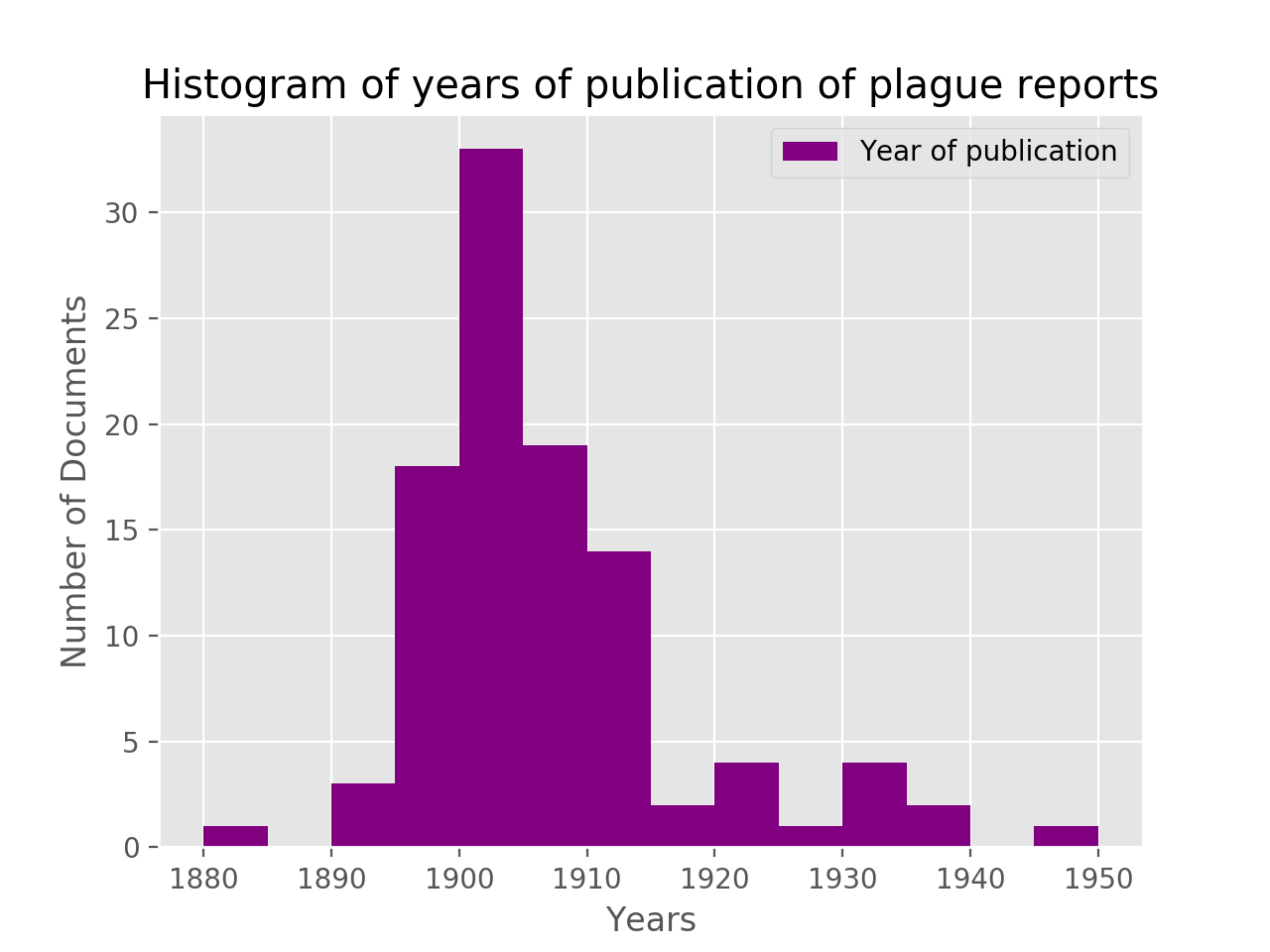}
	\captionof{figure}{Histogram of years of publication for the 102 English reports in the collection.}
	\label{fig:pubyears}
 	\end{centering}
 \end{figure}

The years of publication of English reports in the collection are visualised in the histogram shown in Figure~\ref{fig:pubyears} grouped by 5-year intervals.  The majority of reports were published a few years after the plague pandemic started between 1895 and 1915. A few more reports were published during the tail-end of the pandemic leading up to 1950.  The pandemic was not officially declared over by the World Health Organisation until 1960 when the number of cases dropped below 200 worldwide. However, our collection does not include any reports beyond 1950 as after then there were no major significant outbreaks.

 \begin{figure}[ht]
 \begin{centering}
		\includegraphics[width=0.95\textwidth]{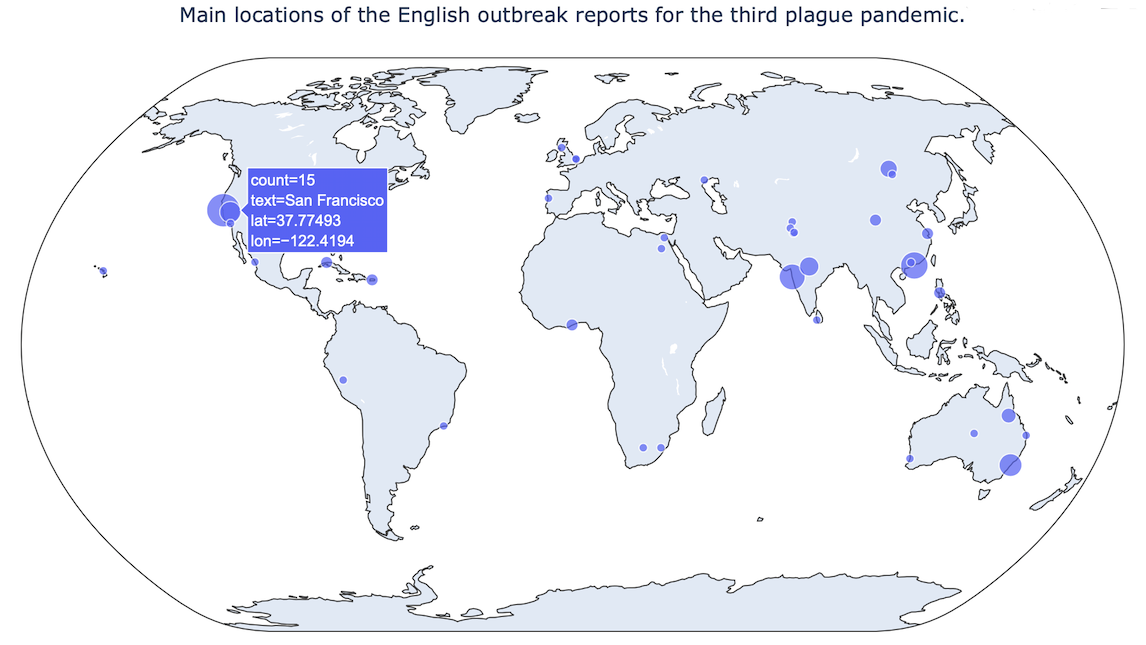}
	\captionof{figure}{Geographical scatter plot of the main outbreak locations of the English reports in the dataset.}
	\label{fig:locs}
 	\end{centering}
 \end{figure}
 
The main locations of the outbreaks described in the reports are visualised in Figure~\ref{fig:locs}.  The size of each mapped location corresponds to the number of reports covering it.  Most reports are about San Francisco, Hong Kong and Bombay but there is a long tail of less frequently covered locations.  The map shows that many of them are located along the coast, cities with ports where the plague spread particularly easily as a result of ongoing trade at that time. Some locations are inland and correspond to country or region names with corresponding latitude and longitude coordinates retrieved from GeoNames.\footnote{\url{https://www.geonames.org}}

\begin{table}[!htbp]
  \newcolumntype{+}{>{\global\let\currentrowstyle\relax}}
  \newcolumntype{^}{>{\currentrowstyle}}
  \newcommand{\rowstyle}[1]{\gdef\currentrowstyle{#1}%
    #1\ignorespaces
  }
\centering
\begin{tabular}{lrrrrr}
\hline
Counts & Total & Min & Max & Mean & Stddev \\\hline
Sentences & 229,043 & 32 & 17,635 & 2,245.5 & 3,713.6\\
Words & 4,443,485 & 1,091 & 396,898 & 43,563.6 & 74,621.0 \\
\hline
\end{tabular}
\caption{Number of sentences and words in the collection of English plague reports, as well as corresponding counts for the smallest document (Min) and the largest document (Max), the average (Mean) and standard deviation (Stddev).
\label{fig:collectionstats}}
\end{table}

Table~\ref{fig:collectionstats} provides an overview of counts of sentences and words in the collection and illustrates the variety of documents in this data.  To derive these counts we used automatic tokenisation and sentence detection over the raw OCR output which is part of the text mining pipeline described in section \ref{sect:annotation}.  While the smallest document is only 32 sentences long containing 1,091 word tokens, the largest report contains almost 400,000 word tokens. The collection contains 38 documents with up to 5,000 words each, 15 reports with between 5,000 and 10,000 words each, 32 documents with between 10,000 and 100K words each and 17 documents with 100K or more words each.  In total, the reports amount to over 4.4 million word tokens and over 229,000 sentences.

Exact details on what articles or works are part of this collection and accessible download links to their pdfs (if available) are provided on the project's GitHub repository.\footnote{\url{https://github.com/Edinburgh-LTG/PlagueDotTxt}}

\subsection{OCR Improvements}
\label{sec:ocr}
\vspace{1ex}

When initially inspecting this digitised historical data, we realised that some of the OCR was of inadequate quality. We therefore spent time during the first part of the project on improving the OCR quality of the reports.

\begin{figure}[ht]
 \begin{centering}
		\includegraphics[width=.8\textwidth]{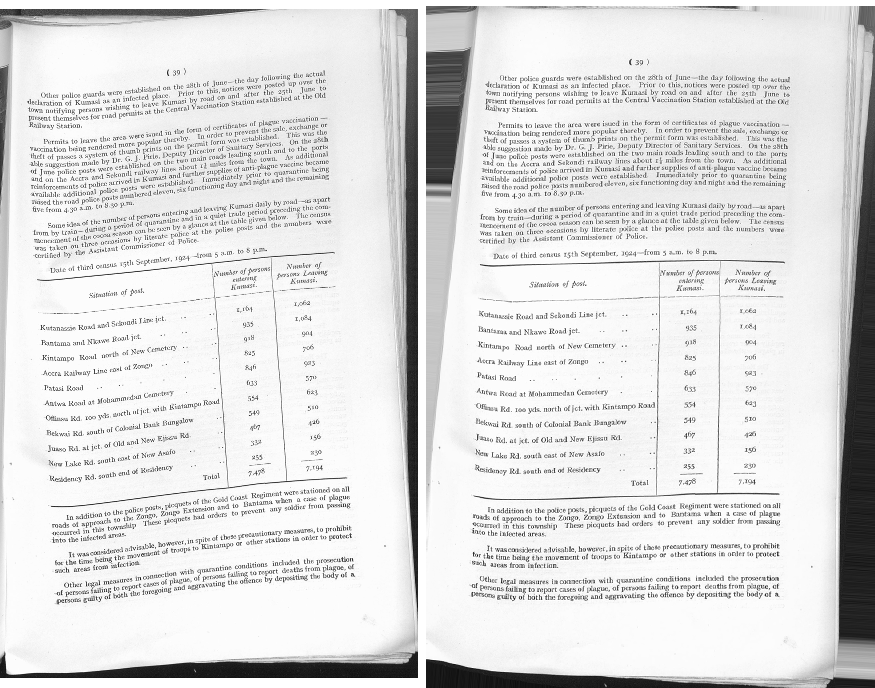}
	\captionof{figure}{A sample page from one of the reports before and after the dewarping process.}
	\label{fig:dewarp}
 	\end{centering}
 \end{figure}

Using computer vision techniques, we processed the report images to remove warping artefacts \citep{Fu-Dewarping}. This was done using Python and the numpy,\footnote{\url{https://numpy.org}} SciPy,\footnote{\url{https://www.scipy.org}} and OpenCV\footnote{\url{https://opencv.org}} libraries. We find the text within each image by binarising and thresholding the image,\footnote{We applied an adaptive 65\% threshold which helped to preserve the text on the page and remove blemishes and text bleeding from the printing on the reverse of a page.} followed by horizontal dilation to connect adjacent letters. Following this, principal component analysis was used to determine the location of text lines in the image, and then OpenCV is used to estimate the ``pose" of the page and generate a reprojection matrix, which is optimised with SciPy using the Powell solver, an optimisation algorithm available in this library.\footnote{\url{https://docs.scipy.org/doc/scipy/reference/optimize.minimize-powell.html}}  An example page image before and after dewarping is shown in Figure~\ref{fig:dewarp}.

\begin{figure}[!htbp]
 \begin{centering}
 	\includegraphics[width=0.75\textwidth]{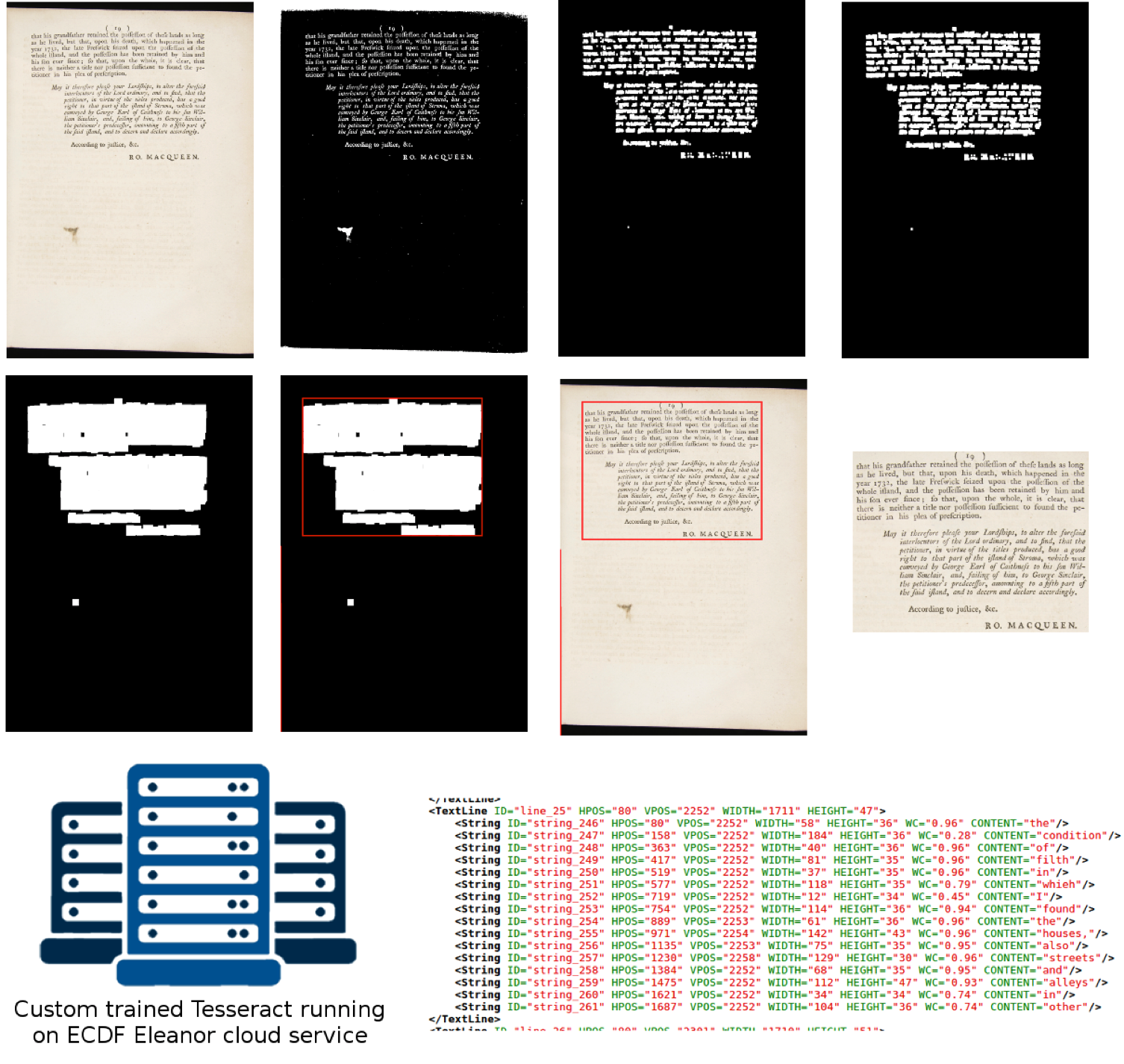}
 	\caption{The OCR process.}
 	\label{fig:ocr-process}
 	\end{centering}
 \end{figure}

We then identified likely textual areas in report images, and produced an effective crop, to provide the OCR engine with less extraneous data (see Figure~\ref{fig:ocr-process}).  This was done with similar methods to the page dewarping, again utilising the OpenCV library to binarise, threshold and dilate the text components of the image. This process was repeated until a maximum target number of contours were present in the image, and then a subset-sum was used to find the most efficient crop.  More information on the methods used and steps taken can be found in the University of Edinburgh Library Labs blog post.\footnote{\url{http://libraryblogs.is.ed.ac.uk/librarylabs/2017/06/23/automated-item-data-extraction/}} OCR was then performed using Tesseract,\footnote{\url{https://opensource.google.com/projects/tesseract}} trained specifically for typeface styles and document layouts common to the time period of the reports.

Training was done across a range of truth data, covering period documents obtained from the IMPACT Project data sets,\footnote{\url{https://www.digitisation.eu/tools-resources/image-and-ground-truth-resources/}} documents from Project Gutenberg prepared for OCR training,\footnote{\url{https://github.com/PedroBarcha/old-books-dataset}} internal ground-truth data compiled as part of the Scottish Session Papers project at the University of Edinburgh\footnote{\url{https://www.projects.ed.ac.uk/project/luc020/brief/overview}} and typeface data sets designed for Digital Humanities collections.\footnote{\url{https://github.com/jbest/typeface-corpus}}

 \begin{figure}[!htbp]
 \begin{centering}
 	\includegraphics[width=0.75\textwidth]{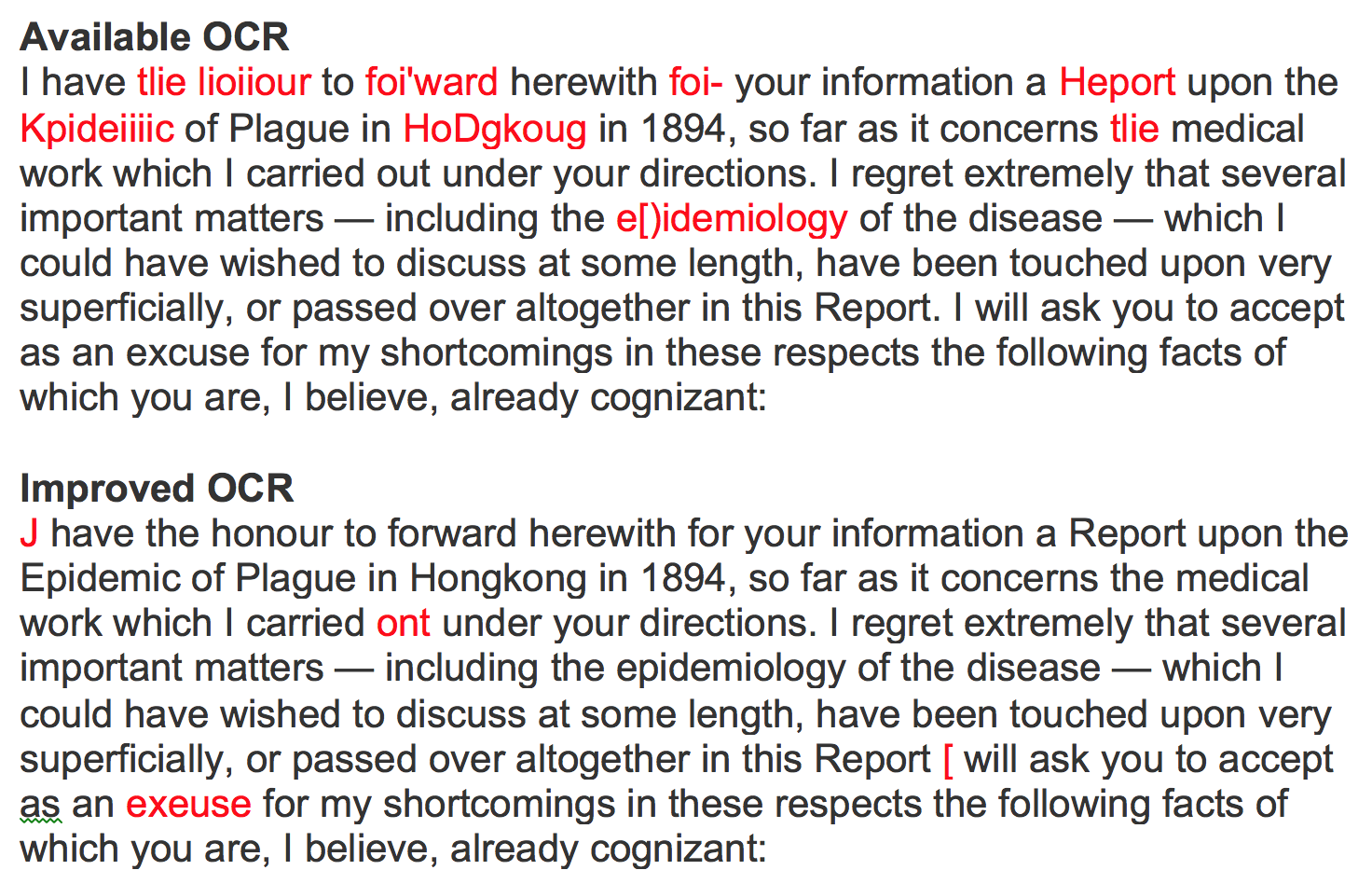}
 	\caption{Excerpt from the Hong Kong report with different versions of OCR output. The Internet Archive image containing this excerpt can be accessed here: \url{https://archive.org/details/b24398287/page/n7}}
 	\label{fig:ocr-example-comparison}
 	\end{centering}
 \end{figure}
 
 \begin{figure}[!htbp]
 \begin{centering}
 	\includegraphics[width=0.75\textwidth]{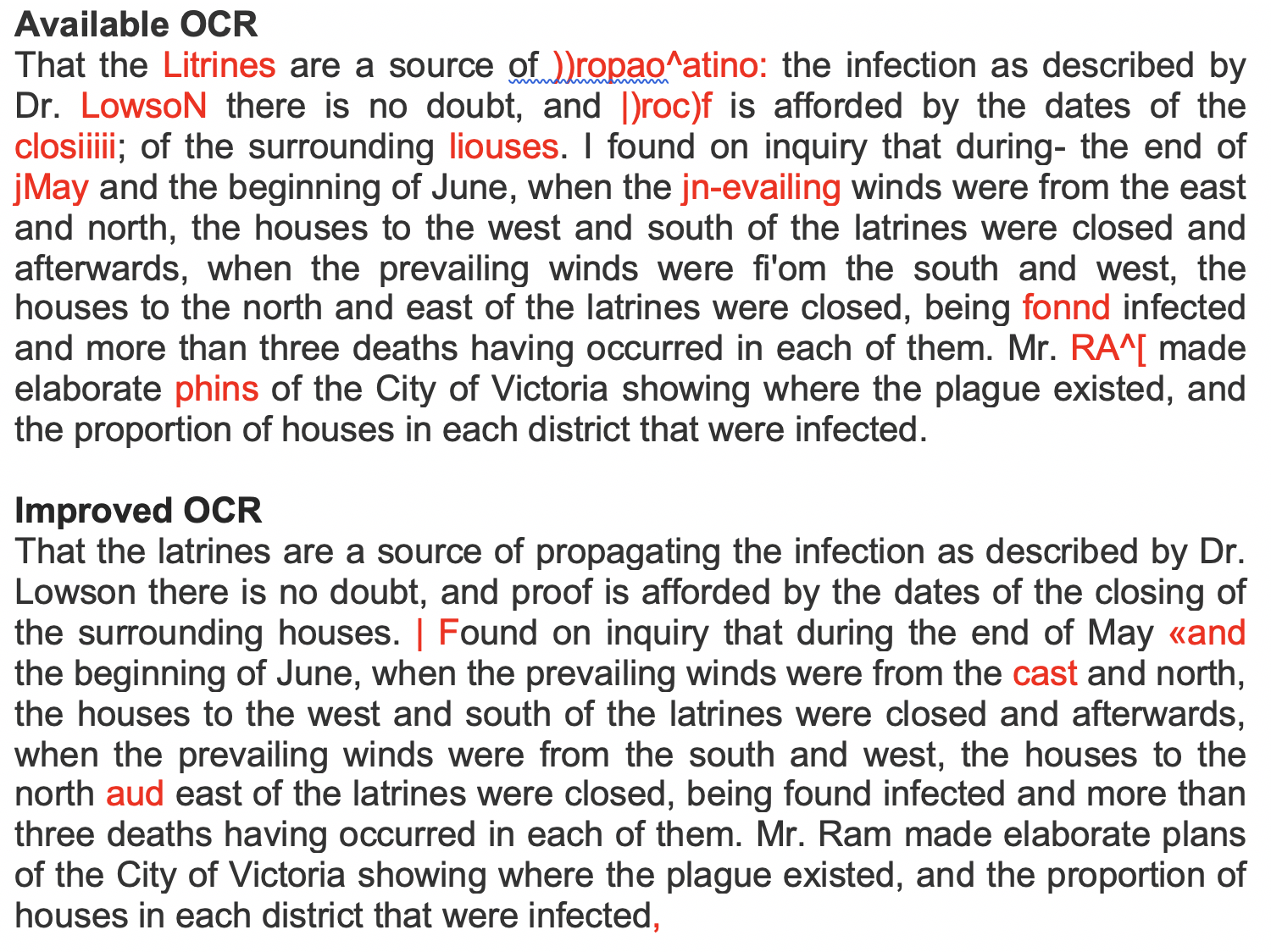}
 	\caption{Another excerpt from the Hong Kong report with different versions of OCR output. The Internet Archive image containing this excerpt can be accessed here: \url{https://archive.org/details/b24398287/page/n7}}
 	\label{fig:ocr-example-comparison2}
 	\end{centering}
 \end{figure}
 
While we have not yet formally evaluated the improvements made to the OCR, observation of the new OCR output shows clear improvements in text quality.\footnote{Previous work conducted by members of the IMPACT project has formally compared OCR quality of ABBYY FineReader and Tesseract \citep{helinski2012report} for different types of test sets.  They have shown that the latter performs more accurately on gothic type documents in terms of both character and word level accuracy.} This is important as it affects the quality of downstream text mining steps.  Previous research and experiments have found that errors in OCRed text have a negative cascading effect on natural language processing or information retrieval tasks \citep{Hauser:2007, Lopresti:2008b, Gotscharek:2011, Alex:2014}. In future work, we would like to conduct a formal evaluation comparing the two versions of OCRed text to quantify the quality improvement.

Figures~\ref{fig:ocr-example-comparison} and \ref{fig:ocr-example-comparison2} shows a comparison of the OCR for two excerpts from the Hong Kong plague report referred to earlier \citep{lowson1895epidemic}.  The excerpts marked as ``Available OCR" refer to the version openly accessible on Internet Archive and created using ABBYY FineReader 11.0.\footnote{\url{https://www.abbyy.com/media/2761/abbyy-finereader-11-users-guide.pdf}} The improved OCR was created using Tesseract as part of the work presented in this paper.  Errors in the OCR are highlighted in red. While a thorough evaluation across the OCRed reports in the corpus is needed to provide a quantitative comparison of OCR quality for both methods, these example excerpts illustrate the types of errors created by them.  Initial observations suggest that the first method appears to struggle to recognise common words like \textit{honour} or \textit{latrines} and names like \textit{Hongkong} and \textit{Dr. Lowson} correctly.  The Tesseract model appears to be more robust towards names and common words in these examples but, in contrast, makes mistakes for the personal pronoun \textit{I} and function words like \textit{and} or \textit{out}. As our analysis is primarily focused on content words, observing the output led us to choose the results produced by the Tesseract model for further processing and annotation. 

\section{Annotation}
\label{sect:annotation}
\vspace{1ex}

This section describes the schema we implemented to structure the information contained within the reports and automatic and manual annotation applied to our collection of plague reports.  We first processed them using a text mining pipeline which we adapted and enhanced specifically for this data.  The text mining output annotations are then corrected and enriched during a manual annotation phase which is still ongoing.  Each report that has undergone manual annotation is then processed further using automatic geo-resolution and date normalisation.

\begin{table*}
  \newcolumntype{+}{>{\global\let\currentrowstyle\relax}}
  \newcolumntype{^}{>{\currentrowstyle}}
  \newcommand{\rowstyle}[1]{\gdef\currentrowstyle{#1}%
    #1\ignorespaces
  }
\begin{center}
\begin{tabular}{ll}
\hline \textbf{Zones} & \textbf{Description} \\\hline
    Title-matter & Title page \\
     Preface&\multicolumn{1}{p{10cm}}{\raggedright Preface information}\\
    Content-page& Content page information     \\
    Introduction &\multicolumn{1}{p{10cm}}{\raggedright State of the epidemic at the time of the production of the report, summary of key features, evaluation of significance of the epidemic}\\
    Disease history & \multicolumn{1}{p{10cm}}{\raggedright General points on the history of the epidemic, origin of outbreak} \\
    Outbreak history & \multicolumn{1}{p{10cm}}{\raggedright Geographical and chronological overview of local outbreak.  What happened this place this year}\\
    Local conditions & \multicolumn{1}{p{10cm}}{\raggedright Descriptions of key elements that are considered noteworthy, something that has contributed or impacted the outbreak}\\
    Causes & \multicolumn{1}{p{10cm}}{\raggedright Causes identified by the author e.g.~usually points of origin, specific local conditions or descriptions of import}\\
    Measures & \multicolumn{1}{p{10cm}}{\raggedright  List of the measures e.g.~undertaken to curb the outbreak, sanitary improvements, 
quarantines, disinfection or fumigation and rat catching}\\
    Clinical appearances &\multicolumn{1}{p{10cm}}{\raggedright Description of the disease appearance, its usual course and its mortality } \\
    Laboratory & \multicolumn{1}{p{10cm}}{\raggedright Description of bacteriological analysis, human lab work}\\
    Treatment & \multicolumn{1}{p{10cm}}{\raggedright Description of the treatment given to patients}\\
    Cases & \multicolumn{1}{p{10cm}}{\raggedright List of individual cases, usually with age, gender, occupation, course of disease, and time and dates of infection and death}\\
    Statistics & \multicolumn{1}{p{10cm}}{\raggedright Contains many lists or tables of statistics such as deaths}\\
    Epizootics & \multicolumn{1}{p{10cm}}{\raggedright Contains information solely about animals, experiments or discussions} \\
    Appendix & Labelled appendix\\
    Conclusion& Conclusion\\
  \hline
  \end{tabular}
  \end{center}
  \caption{Zones
  \label{tab:zones}}
\end{table*}

\subsection{Developing an Annotation Schema}
\label{sect:approach}
\vspace{1ex}

As discussed in the Background section, our methodological approach is based on genre analysis \citep{Swales:1990uv,Bhatia}  which treats each report as a communicative event.  We hypothesise that the reports - despite their variation of styles - will present and structure their arguments comparably, as they are intended for the same audience of fellow epidemiologists and government officials.  Thus we assume to find comparative segments of text which discuss a similar theme e.g. \textit{measures taken, local conditions} across the collection of reports.  We refer to these comparative segments of text as zones where each zone has a specific purpose, described in Table \ref{tab:zones}. Within each zone, the author uses the narrative to build an argument or convey thinking about the zone's theme.  For example, within a \textit{measures} zone,  authors have discussed measures taken to prevent the spread of the disease and their impact. 
The collation of report narratives into zones is not straightforward as authors approach the narrative with different styles and label text with different titles. For example, one may call a section of text \textit{Background} and another may call it \textit{Outbreak History} making the application of a schema to support automated labelling challenging. Therefore labelling of zones is done manually by annotating text through close reading of the report. However,  in the future we intend to investigate if this could be approached in an automated way. The list of zones we have chosen as a scheme for annotation has emerged both from the formal conventions of published reports (with regards to the report's apparatus, containing \textit{title-matter, preface, footnotes}) as well as from extensive historical research. Zones that emerged from sections and chapters within some  reports were aligned with overarching concepts and categories, which epidemiologists used at the time.  

In addition to our zoning schema, we also annotated for a number of entities within the text (Table \ref{tab:entities}).  This supports the comparative analysis across zones allowing entities to be tracked, such as \textit{location}, \textit{plague term}, \textit{date} and \textit{time}.  The zoning schema and entity list was created from studying a subsection of reports, section titles and three rounds of pilot annotation on a subsection of documents.

\subsection{Automatic Annotation and Text Mining}
\label{sect:ie}
\vspace{1ex}

\begin{figure}[!htbp]
    \centering
 	\includegraphics[width=0.8\textwidth]{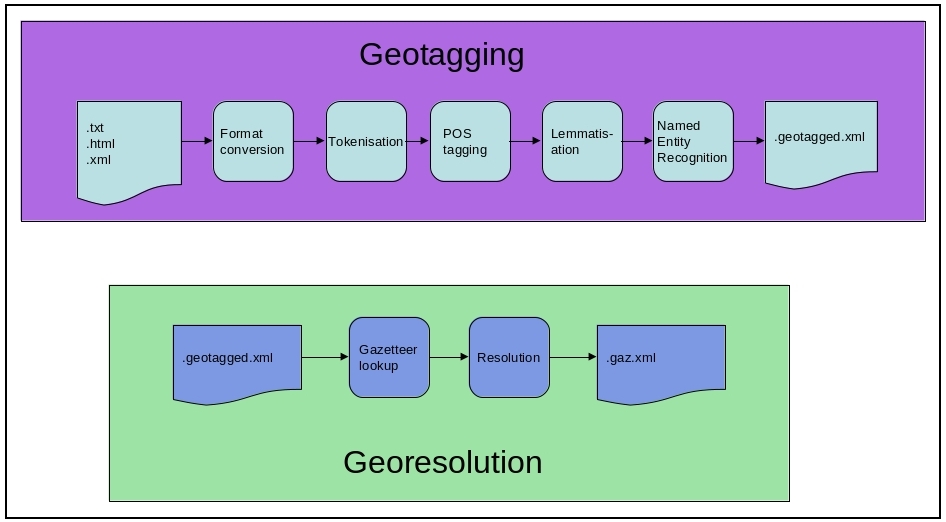}
 	\caption{The Edinburgh Geoparser pipeline.}
 	\label{fig:geoparser}
\end{figure}

To process the plague reports, we used the Edinburgh Geoparser \citep{Grover:2010}, a text mining pipeline which has been previously applied to other types of historical text \citep[for example]{Rupp:2013,Alex:2015b,Clifford:2016,Rayson:2017, Alex:2019}.  This tool is made up of a series of processing components. It takes as input raw text and performs standard text pre-processing on documents in XML format, including tokenisation, sentence detection, lemmatisation, part-of-speech tagging and chunking as well as named entity recognition and entity normalisation of dates and geo-resolution in the case of location names (see Figure~\ref{fig:geoparser}).  The processing steps are applied using LT-XML2, our in-house XML tools~\citep{Grover2006}.\footnote{\url{https://www.ltg.ed.ac.uk/software/ltxml2/}} Before tokenisation, we also run a script to repair broken words which were split in the input text as a result of end-of-line hyphenation described in \citep{Alex:2012}.

\begin{table}
  \newcolumntype{+}{>{\global\let\currentrowstyle\relax}}
  \newcolumntype{^}{>{\currentrowstyle}}
  \newcommand{\rowstyle}[1]{\gdef\currentrowstyle{#1}%
    #1\ignorespaces
  }
\centering
  \begin{tabular}{ll}
    \hline
    Entity Type & Entity Mentions \\\hline
    person & Professor Zabolotny, Professor Kitasato, Dr. Yersin, M. Haffkine \\
    location & India, Bombay, City of Bombay, San Francisco, Venice\\
    geographic-feature & house, hospital, port, store, street\\
    plague-ontology-term & plague, bubo, bacilli, pneumonia, hemorrhages, vomiting\\
    date & 1898, March 1897, 4th February 1897, the beginning of June, next day\\
    date-range & 1900-1907, July 1898 to March 1899, since September 1896 \\
    time & midnight, noon, 8 a.m., 4:30 p.m.\\
    duration & ten days, months, a week, 48 hours, winter, a long time \\
    distance & 20 miles, 100 yards, six miles, 30 feet \\
    population/group of people & Chinese, Europeans, Indian, Russian, Asiatics, coolies, villagers\\
    percent & 8\%, 25 per cent, ten per cent\\
   \hline
  \end{tabular}
  \caption{Entity types and examples of entity mentions in the plague reports.
  \label{tab:entities}}
\end{table}

We adapted the Edinburgh Geoparser by expanding the list of types of entities it recognises in text, including geographic-feature, plague-ontology-term and population/group of people etc.\footnote{Note that our goal was to emulate descriptions used by the authors at the time, mirroring concepts of race and ethnicity that were often implicated in the construction of epidemiological arguments. Some examples of the population/group of people annotations show that these are often derogatory and considered offensive today. They are strictly understood to be of value only for the illumination of historical discourse.}  A  list of entity types extracted from the plague reports and examples are presented in Table~\ref{tab:entities}.  Date entity normalisation and geo-resolution provided by the default Edinburgh Geoparser are re-applied once the manual annotation (described in the next section) for a document is completed.  This is to ensure that the corrected text mining output is geo-resolved and normalised correctly for dates, including manual corrections of spelling mistakes occurring in entity mentions.

The main effort in adapting the Edinburgh Geoparser was directed towards adding additional entity types to those recognised by the default version (e.g. geographic-feature, plague-ontology-term, and population).  Plague related terminology (plague-ontology-term) is recognised using a domain-specific lexicon of terms relevant to the third plague pandemic.  This was bootstrapped using manual annotation of plague terms in the pilot phase (including ones containing OCR errors) and extending this list by allowing matches of different forms (singular/plural) and crucially adding manually corrections of OCR errors as attributes to the entity annotations. This enables us to add annotations automatically to text still containing OCR errors with the aim to do further OCR post-correction or allow keyword searching over text containing these errors and including them in the results even if the search term is typed correctly. Geographic features are marked up using similar lexicon matching which is complemented by adding further geographical features that are derived automatically using WordNet,\footnote{\url{https://wordnet.princeton.edu}} a lexical database of semantic relations.  The latter approach is also used to recognise population entities.

\subsection{Manual Annotation}
\label{sect:manualannot}
\vspace{1ex}

The bulk of the manual annotation has been carried out by two main annotators, a PhD student trained in natural language processing and a medical anthropologist PhD student.  Prior to the main annotation phase we conducted a pilot annotation which lasted one week in order to train both annotators how to use the annotation tool, what information to mark up and to refine details in the annotation guidelines.  This pilot annotation involved direct input and feedback from the academics leading this project (a computational linguist and a medical historian).  After the pilot, the two annotators started annotating the data independently but asked any queries they had to the group.  The balance of historians and NLP experts on this project worked as an advantage as the former bring the knowledge about the data, the historical background and ideas of what information is needing to be captured in the annotation and the latter have the expertise in the technology and methods used when applying natural language processing to automate or semi-automate some of the steps in this process. 

Manual annotation was necessary for a number of reasons. Whilst some zones could be identified automatically from section titles we found that this was often hampered by spelling errors due to OCR issues arising from typeface styles and title placements in margins.  In addition, depending on author narrative styles some zones could be found nested within sections with no titles. This created the need to manually annotate zones.  The automatic recognition of named entities (see Section~\ref{sect:ie}) was partially successful but also suffered from spelling errors and OCR issues.  In addition, as reports were annotated new entity mentions were identified. Thus manual additions or correction of erroneous or spurious entity mentions was deemed necessary.

\begin{figure}[ht]
    \centering
	\includegraphics[width=0.99\textwidth]{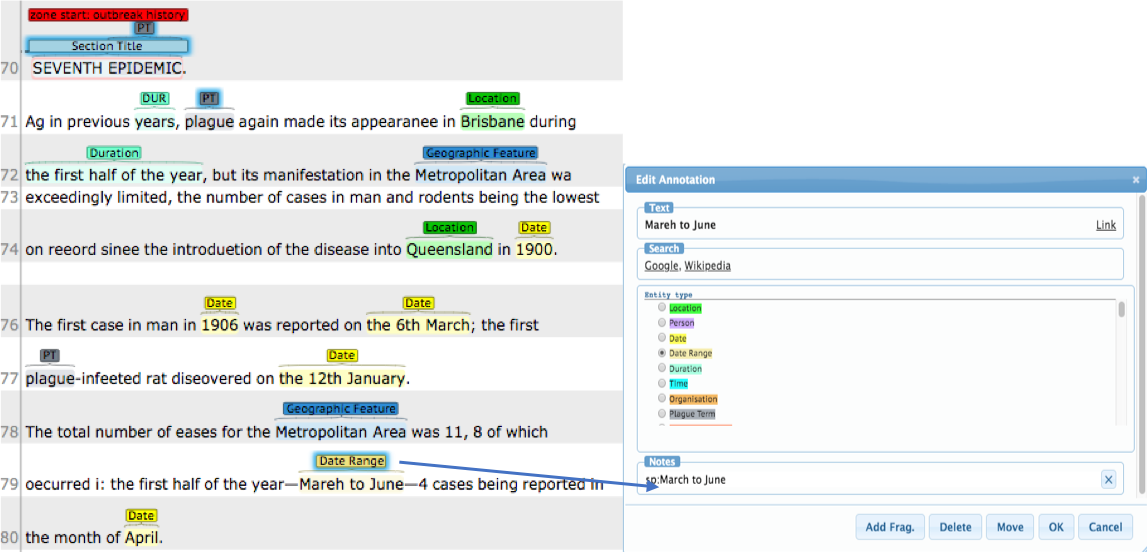}
	\caption{Brat annotation tool.}
	\label{fig:brat}
\end{figure}

Manual annotation is conducted using Brat,\footnote{\url{ https://brat.nlplab.org/}} a web-based text annotation tool \citep{Stenetorp:2012}. After the text was processed automatically as described above, it was converted from XML into Brat format to be able to correct the text mining output and add zone annotations.\footnote{We have not yet conducted double annotation to determine inter-annotator agreement for this work but this is something we are planning to do in the future.} Figure~\ref{fig:brat} shows an excerpt of an example report being annotated in Brat.  Entities such as date, location or geographic feature listed in Table~\ref{tab:entities} can be seen highlighted in the text.  The start of an \textit{outbreak history} zone is also marked at the beginning of the excerpt.

\subsubsection{Zone Annotation}
\vspace{1ex}

Zone annotation, as defined by our schema shown in Table~\ref{tab:zones}, is applied inclusive of a section title and can be nested.  For example, zones of \textit{cases} are often found inside \textit{treatment} or \textit{clinical appearances} zones.  \textit{Footnote} zones were added as these often break the flow of the text and make downstream natural language processing challenging.  In addition, we added \textit{Header/Footer} markup to be able to exclude headers and footers, e.g.~the publisher name or title or section title of a report repeated on each page, from further analysis or search.

Tables were a challenge for the OCR and unusable for the most part.  When marking up tables, we also record their page number.  Text within tables is currently ignored when ingesting the structured data to Solr (see Section~\ref{sect:interfaces}). However, tables include a lot of valuable statistical information. In the next phase of the project we will investigate whether this information can be successfully extracted or whether it will need to be manually collated.

\subsubsection{Entity Annotation}
\vspace{1ex}

During manual annotation we instruct our annotators to correct any wrongly automated entities and add those that were missed.  Any mis-spellings of entity mentions, mostly caused by the OCR process, are also corrected in the Note field in Brat. An example date-range annotation containing an OCR error, \textit{Mareh to June} corrected to \textit{March to June}, is shown in Figure \ref{fig:brat}.  The mis-spellings are subsequently used as part of our text cleaning process.  The corrected forms are also used to geo-resolve place names and normalise dates. These final two processing steps of the Edinburgh Geoparser are carried out on each report once it has been manually annotated and converted back to XML.

\section{Data Search Interface}
\label{sect:interfaces}
\vspace{1ex}

One goal of the \textit{Plague.TXT} project is to make our digital collection available as an online search and retrieval resource but in addition this collection should be accessible. This means being available for example, to computational linguists as an annotated resource for direct in-depth analysis as well as via interactive search for humanities researchers.  This provides vital support for historians and humanities researchers improving on the limited capacities of manual searches through document collections to find information pertinent to their research interest.  Additionally, the challenges of working with such text digitally require interdisciplinary collaboration. HistSearch \citep{petterson:2016}, an on-line tool applied to historical texts, demonstrates how computational linguists and historians can work together to automate access to information extraction and we will evaluate similar approaches for this collection.  We plan to make the digital collection available with Apache Solr.\footnote{\url{https://lucene.apache.org/solr/}\newline \indent See this website for a further description of Solr features.}

Solr is an open-source enterprise-search platform, widely used for digital collections.  The features available through the Solr search interface make our collection accessible to a wide audience with varying research interests. It offers features that support grouping and organising data in multiple ways, while data interrogation can be achieved through its simple interface with term, query, range and data faceting.  Solr also supports rich document handling with text analytic features and direct access to data in a variety of formats.

We are currently customising and improving the filtering of the data for downstream analysis in Solr. In the following section, we describe on-going filtering steps with Solr and provide examples to demonstrate a search interface customisation. Further, we explore preliminary analysis that can be done from data retrieved via the search interface.

\subsection{Data Preparation and Filtering in Solr}
\label{sect:dataprep}
\vspace{1ex}

The annotated data is prepared and imported to Solr using Python, with annotations created both automatically by the Geoparser and manually by the annotators mapped to appropriate data fields (e.g.~date-range entities are mapped to a Date Range field,\footnote{\url{https://lucene.apache.org/solr/guide/8_1/working-with-dates.html\#date-range-formatting}}) enabling complex queries across the values expressed in the document text. Additionally, manual spelling corrections are used to replace the corresponding text in the OCR rendering prior to Solr ingestion, thus improving the accuracy of language-based queries and further textual analysis.  We also implement lexicon-based entity recognition for entities that have been missed during the annotation and for additional entity types, e.g animals.  Solr allows for storing and searching by geo-spatial coordinates and we import geo-coordinates associated with entities identified by the Geoparser.  Geo-coordinates can be used to support interactive visualisations, as developed in the Trading Consequences project \citep{hinrichs2015} which visualises commodities through their geo-spatial history. In addition, this location information can be used in analysis such as transmission and spread, e.g.~geo-referenced plague outbreak records have been used to show how major trade routes contributed to the spread of the plague \citep{Yue:2017}.   Using \textit{case} zones we are currently assessing NLP techniques to extract case information into a more structured format for direct access to statistical information from hundreds of individual case descriptions.

\section{Use cases for the Plague.TXT Data}
\vspace{1ex}

Historians and computational linguists have different methods and reasons for analysing a data set.  The Plague.TXT team not only provide a search and exploration interface but will also release the underlying data (for titles with permissible licenses) to allow direct corpus analysis.  In this section, we provide examples for three different use cases of this data.

\subsection{Use Case 1: Illustration of Interactive Search}
\vspace{1ex}

\begin{figure}[!htbp]
    \centering
 	\includegraphics[scale=0.8]{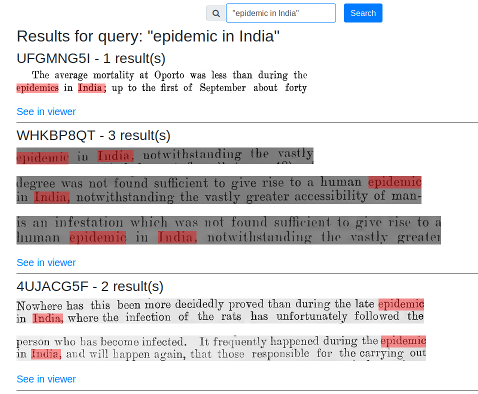}
 	\caption{Solr snippet search example.}
 	\label{fig:search-snip}
\end{figure}

\begin{figure}[!htbp]
    \centering
 	\includegraphics[scale=0.6]{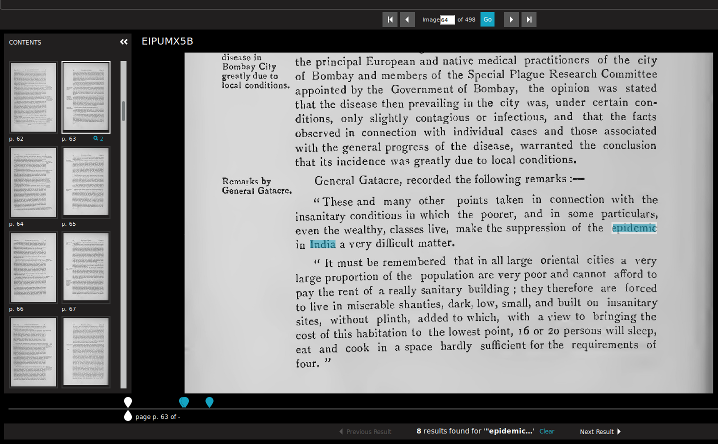}
 	\caption{IIIF viewer search example.}
 	\label{fig:search-viewer}
\end{figure}

Figure~\ref{fig:search-snip} shows one of our customised search interfaces. This allows users to search across the entire report collection displaying original image snippets from the reports containing the search term(s).  This search function enables the user to grasp the immediate context of search-terms within the page and also to recognise potential limits of the OCR recognition currently applied.

Snippet search is supported by indexing OCR transcriptions from word-level ALTO-XML\footnote{\url{http://www.loc.gov/standards/alto/}} in Solr and then by using Whiiif,\footnote{\url{https://github.com/mbennett-uoe/whiiif}} an implementation of the International Image Interoperability Framework (IIIF) Search API\footnote{\url{https://iiif.io/api/search/1.0/}} designed to provide full-text search with granular, word-level annotation results to enable front-end highlighting.

Figure~\ref{fig:search-viewer} shows a similar search within a single document using UniversalViewer,\footnote{\url{https://universalviewer.io}} a IIIF viewer utility. This document-level search is powered by the same Whiiif\footnote{Whiiif stands for Word Highlighting for IIIF. Further technical details about the Whiiif package can be found on the University of Edinburgh Library Labs blog: \url{http://libraryblogs.is.ed.ac.uk/librarylabs/2019/07/03/introducing-whiiif/}} instance, again making use of the IIIF Search API to provide a method of in-document searching that is available natively within any compatible IIIF viewing software. This functionality is made available to visualise the search results within the document context, as part of a whole-document browsing interface, allowing for greater context of the search results to be shown.

 \subsection{Use Case 2: Finding Discussion Concepts in Causes Across Time Periods}
\vspace{1ex}

Our search interfaces facilitate queries across the collection on content and meta-data as well as queries based on zones or entity types using facets such as date range.  In this use case we focus on topics discussed in \textit{cause} zones and if these differ between report time periods. We do this through extracting the data via Solr and applying topic modelling.  First, we search for \textit{cause} zones published during the pandemic 1894-6 comparing these to \textit{cause} zones in reports 1904 and beyond.  We retrieve the results via the Solr API in XML format and apply removal of stop-words and all non dictionary terms to the \textit{cause} zone text. In future, we will make indexed versions of cleaned data in this format directly accessible from Solr.  We use topic modelling (LDA with Gensim Python library\footnote{\url{https://radimrehurek.com/gensim/index.html}}) to compare the \textit{cause} zone text at the different time points, selecting two topics.  

\begin{table}[!htbp]
  \newcolumntype{+}{>{\global\let\currentrowstyle\relax}}
  \newcolumntype{^}{>{\currentrowstyle}}
  \newcommand{\rowstyle}[1]{\gdef\currentrowstyle{#1}%
    #1\ignorespaces
  }
\centering
\begin{tabular}{ll}
\hline
    Topic/Date &  keywords\\\hline
    \textbf{(1) 1894-96} &  \multicolumn{1}{p{5cm}}{\raggedright latrine, house, soil, street, find, case, time, plague,
    infection, opinion, condition, may, must, question, see}\\
    \textbf{(2) 1894-96} &  \multicolumn{1}{p{5cm}}{\raggedright
    house, people, ordinance, well, supply, cause, must, condition,
    drain, disease, pig, matter, area, water, provision}\\ \hline
   \textbf{(1)  1904-07}&  \multicolumn{1}{p{5cm}}{\raggedright
    plague, rat, case, infection, man, flea, may, infect, place,
    fact, evidence, disease, instance, produce, find}\\
    \textbf{(2) 1904-07} &  \multicolumn{1}{p{5cm}}{\raggedright year, month, temperature, epidemic, influence, season, infection,
    december, condition, may, june, prevalence, rat, follow, number}\\
   \hline
  \end{tabular}
  \captionof{table}{Discussion topics from cause zones by time period
  \label{tab:topics}}
\end{table}

Results are presented in Table \ref{tab:topics}.  The earlier reports show the discussion centering around environment aspects with focus on populations, conditions of living and buildings and how this might cause the spread.  The second topic is linked to the concepts at that time period, about how the diseases may spread through the water system, with studies of ordinance maps of sewerage and water ways. Looking at the later reports we now see rats and fleas and infection are more prominent as a discussion topic but also season, temperature and weather form a topic being discussed as a causal factor. Combined further with geo-resolution information, this type of zone-specific topic analysis across time periods could reveal interesting patterns and inferences about the reasoning  of epidemiologists observing outbreaks.

\subsection{Use Case 3: Corpus Analytics}
\vspace{1ex}

Analysing a corpus with respect to token frequencies can reveal interesting patterns and insights into aspects such as gender, age and population.  In this use case we look at the corpus with respect to gender and consider how men and women are mentioned within the corpus.
As all reports in the collection are tokenised and part-of-speech tagged, frequency-based and syntactic-category-dependent corpus analysis can be conducted across the collection.\footnote{A syntactic category corresponds to a part of speech of a text token (e.g.~noun, verb, preposition, etc.).  Syntactic-category dependent corpus analysis is therefore counting tokens that are tagged with a particular part of speech tag.} The ratio of the total number of mentions of \textit{woman} or \textit{women} versus \textit{man} or \textit{men} is 0.19 (681 versus 3603 mentions after lower-casing the text).  Similarly the ratio for the pronouns \textit{she} versus \textit{he} is 0.15 (1233 versus 8008 mentions after lower-casing).  

Table~\ref{tab:corpus-analysis} lists the twenty most frequent adjectives followed by \textit{man/men} versus \textit{woman/women}.  For the majority of mentions, men are described as \textit{medical}, \textit{young} and \textit{sick} and women as \textit{old}, \textit{married} and \textit{pregnant}.  A similar analysis for verbs following pronouns, the most frequent verbs following \textit{he} (excluding \textit{has}, \textit{is}, \textit{could}, \textit{would} etc.) are \textit{thought} (n=119), \textit{died} (n=80) and \textit{found} (n=65). The phrase \textit{she died}, on the other hand, appears only 15 times out of over 4.2 million words in the collection. This difference is comparable with the ratio of mentions of woman/woman versus man/men (0.19). However, the ratio is much more skewed for the phrases \textit{she thinks/thought} (n=2) versus \textit{he thinks/thought} (n=144).

\begin{table}[!htbp]
  \newcolumntype{+}{>{\global\let\currentrowstyle\relax}}
  \newcolumntype{^}{>{\currentrowstyle}}
  \newcommand{\rowstyle}[1]{\gdef\currentrowstyle{#1}%
    #1\ignorespaces
  }
\centering
\begin{tabular}{rlrl}
\hline
\multicolumn{2}{p{5cm}}{adjective + man$|$men} & \multicolumn{2}{p{5cm}}{adjective + woman$|$women} \\\hline
count & adjective & count & adjective \\\hline
    316 & medical & 21 & old \\
     27 & young & 12 & married \\
     22 & sick & 10 & pregnant \\
     19 & old & 9 & young \\
     13 & medieal & 8 & chinese \\
      9 & poor & 7 & purdah \\
      8 & influential & 5 & native \\
      7 & other & 4 & other \\
      7 & infected & 3 & parturient \\
      7 & healthy & 3 & dead \\
      6 & intelligent & 2 & well-nourished \\
      6 & few & 2 & unfortunate \\
      5 & well-nourished & 2 & indian \\
      5 & twelve & 2 & few \\
      5 & scientific & 1 & weakly \\
      4 & white & 1 & sick \\
      4 & trained & 1 & several \\
      4 & several & 1 & russian \\
      4 & muscular & 1 & respectable \\
      4 & great & 1 & purdak \\
\hline
  \end{tabular}
  \caption{Most frequent adjectives followed by the nouns \textit{man} or \textit{men} versus \textit{woman} or \textit{women}.}
  \label{tab:corpus-analysis}
\end{table}

While these results are unsurprising given that the reports were written over a century ago and authored by men for men, they do raise questions on gender statics within the reporting of cases in these reports.  More thorough analysis, for example by exploring this text in context of its time (see the DICT method proposed by \citet{jatowt:2019}), combined with close reading is necessary to explore these differences in more detail. The search interface to the collection, however, helps to find instances of these mentions in, for example, the context of \textit{case} zones, and thereby supports navigation of the collection more rapidly.

\section*{Discussion, Conclusions and Future Work}
\label{sect:discussion-etc}
\vspace{1ex}

In this paper we have presented the work undertaken in the pilot stage of our \textit{Plague.TXT} project.  The work is the outcome of an interdisciplinary team working together to understand the nature and complexities of a historical text collection and the needs of the potential different types of users of this collection.  A major contribution of this project is the development of a model to capture the narrative structure of the collection of reports. This brings individual reports together in one collection enabling streamlined and efficient linking of knowledge and themes used in the comprehension of the third plague pandemic, covering the time period of the collection. This approach enables analysis of these reports across sections as one coherent corpus. Making this collection accessible through the Solr search interface, we can share it with the research community in ways that cater for the needs of different field experts

Our work in this project is ongoing as we add more data but manual annotation is time consuming and can be an error prone process. As we increase the number of reports annotated with zone markup, we intend to investigate how the annotation can be automated. Possible solutions include: methods, such as content similarity measures which have been shown to be successful in scientific article recommendation \citep{Heconaware}, or work in identifying clinical note duplication  \citep{10.1197/jamia.M3390} which uses distance between words, or work  that measures similarity of scientific articles using divergence of distributions of words \citep{huang2019holes}.

Currently we are developing methods to directly access the statistical information contained within \textit{case} zones and within tables.  Additionally, we will explore spelling normalisation further, such as diachronic and synchronic spelling variance. As well as the methods for spelling normalisation previously mentioned we will also use fuzzy string matching capabilities within Solr to correct for spelling variation introduced by OCR.    Additionally, we will explore changes in semantics over time and how this may impact search and downstream analysis.

\section*{Acknowledgements}
\vspace{1ex}

This work was funded by the Challenge Investment Fund 2018-19 from the College of Arts, Humanities and Social Sciences, University of Edinburgh.

\label{sect:bib}
\bibliographystyle{plainnat}
\bibliography{references}

\end{document}